\documentclass[twocolumn]{article}
\usepackage[utf8]{inputenc}

\usepackage{csquotes}
\usepackage{dirtytalk}
\usepackage{authblk}
\usepackage{graphicx}
\graphicspath{ {img/} }
\usepackage[dvipsnames]{xcolor}
\usepackage{amsmath}
\usepackage{amsfonts}

\title{Through-life Monitoring of Resource-constrained Systems and Fleets}

\author[1]{Felipe Montana}
\author[1]{Adam Hartwell}
\author[1]{Will Jacobs}
\author[1]{Visakan Kadirkamanathan}
\author[1]{Andrew R Mills}
\author[2]{Tom Clark}
\affil[1]{Department of Automatic Control and Systems Engineering, University of Sheffield, UK}
\affil[2]{Rolls-Royce Plc, UK}
\date{}

\begin{document}
\maketitle


\begin{abstract}

A Digital Twin (DT) is a simulation of a physical system that provides information to make decisions that add economic, social or commercial value. The behaviour of a physical system changes over time, a DT must therefore be continually updated with data from the physical systems to reflect its changing behaviour.  For resource-constrained systems, updating a DT is non-trivial because of challenges such as on-board learning and the off-board data transfer. This paper presents a framework for updating data-driven DTs of resource-constrained systems geared towards system health monitoring. The proposed solution consists of: (1) an on-board system running a light-weight DT allowing the prioritisation and parsimonious transfer of data generated by the physical system; and (2) off-board robust updating of the DT and detection of anomalous behaviours. Two case studies are considered using a production gas turbine engine system to demonstrate the digital representation accuracy for real-world, time-varying physical systems.

\end{abstract}

\section{Introduction}\label{sec:Introduction}

Digital Twin (DT) is an emerging technology that has recently gained attention due to the rapid development of simulations, data acquisition and data communication that trigger interaction between the physical and virtual spaces \cite{tao2018digital}. A DT is a simulation of an as-built system, e.g., a vehicle or factory, that accurately represents its corresponding twin \cite{glaessgen2012digital}. In contrast to a simple model or simulation, a DT is a living and evolving model that follows the lifecycle of its physical twin \cite{barricelli2019survey}. This is done by integrating data collected from the system's sensors, environment, historic maintenance data, and available system knowledge. Moreover, DTs can contain a description of the structure, functions, behaviour and control of the physical system \cite{leng2021digital}. The information provided by the DT facilitates the processes of making decisions that will affect the physical asset \cite{DTHub} and adds economic, social or commercial value to stakeholders. 

The DT paradigm is used in a variety of applications and sectors, including product design and factory optimisation \cite{qi2018digital}, prediction of aircraft structural life \cite{tuegel2011reengineering}, and monitoring of automotive braking systems \cite{magargle2017simulation} among others. Another example is the management and monitoring of a fleet of assets. In this case, information based on the fleet distribution is inadequate to assess individual systems due to the variability in the usage, manufacturing and material properties of different assets \cite{li2017dynamic}. The availability of individually tailored DTs is therefore desirable to accurately predict future performance against requirements and detect deviation from the current system behaviour driven by emerging faults.

The behaviour of a physical system is expected to change over time due to normal degradation, mechanical modifications, etc. The result is a drift between the real system and its twin. Another cause of deviation between a real system and its twin, particularly in data-driven DTs, is operation in conditions previously unseen by the twin. Both scenarios motivate the update of the DT using data acquired from the physical system to provide a better representation of the current state of the system. 

For applications such as predictive maintenance and monitoring, updates to the DT mainly consist of updating model parameters. Different methods for parameter updating have been proposed such as solving an optimisation problem \cite{wang2019digital} or using Gaussian processes regression \cite{chakraborty2021machine}. However, little attention has been paid to the detection of anomalous data during the updating process, despite such data being ubiquitous in real-world applications. Anomalous data should be detected, then removed from the update data set, before the DT update. This is essential to stop the DT adapting itself to represent emerging faults which would otherwise make the detection of anomalous behaviours impossible \cite{fink2020potential}.
 
\begin{figure}[ht]
\includegraphics[width=0.49\textwidth]{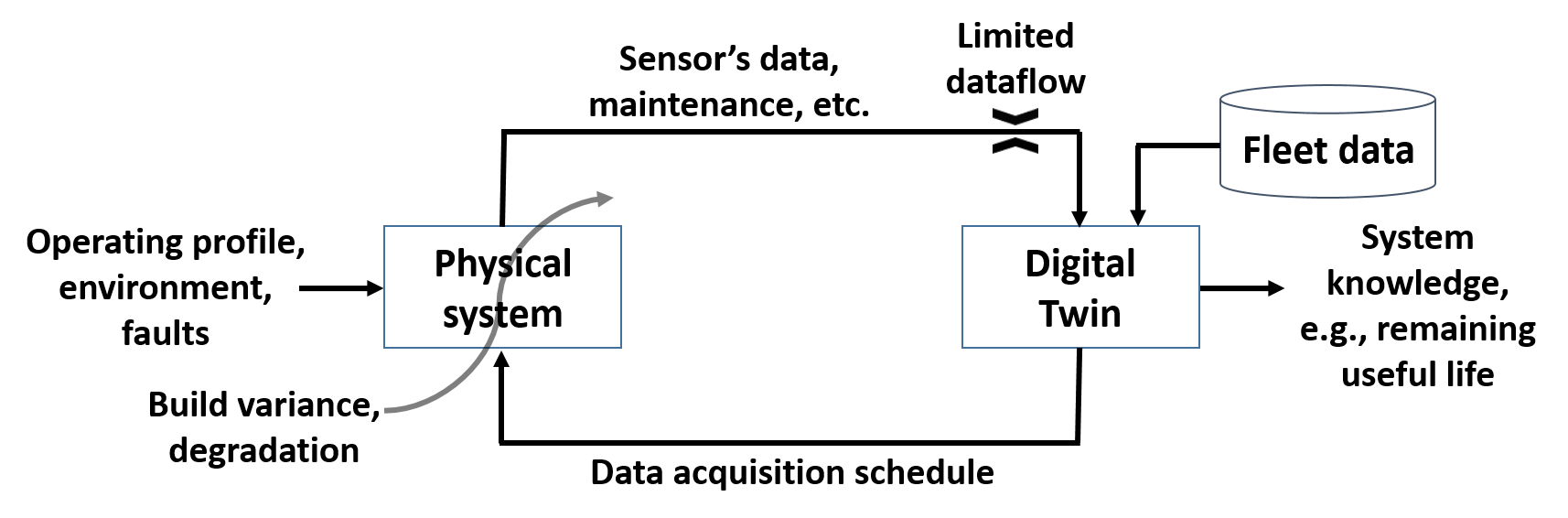}
\caption{\textbf{Data flow in a Digital Twin.} A Digital Twin must be routinely updated, with data recorded from the physical system, to reflect changes due to degradation, build variance, etc. Anomalous data and constraints in data collection are some of the challenges faced when updating the Digital Twin. The Digital Twin provides knowledge about the physical system and the means to simulate the system under different conditions.}
\label{fig:DataFlow}
\end{figure}

The process to update a model is typically many-fold more computationally demanding than its execution. In many real engineering systems, computational power to perform these updates is severely limited ``on-board'', e.g., in the proximity of a gas turbine engine.  Not only is the computational power directly connected to asset limiting for on-line learning, but there is also limited computational storage capacity, low data transmission capacity, or high cost of transmission limits the volume of data that may be transferred for remote updating of the DT. In addition to the data required to update the DT, data containing possible anomalous system behaviour must also be collected for the purpose of health monitoring, e.g., to identify faults in the physical system. There is therefore a joint challenge of continuously monitoring the asset to detect undesirable changes in behaviour, while also updating the model to reflect expected or nominal changes in behaviour, these challenges are illustrated in Fig. \ref{fig:DataFlow}.


To address these challenges, we propose a data-driven approach to develop DTs for groups of computationally resource-limited systems. We limit the scope of the DT to applications such as predictive maintenance and system monitoring. Nevertheless, the updating framework is applicable to other applications of DT. Physical systems, which exhibit non-linear dynamic behaviour, are modelled using a deep neural network trained with data from each individual asset. In contrast to model-based methods that require physical knowledge of the system, data-driven DTs only rely on data to accurately represent their physical twins. Therefore, our framework can be used to represent a variety complex systems. To update the DT over time, segments of data that are not well understood, i.e., are predicted with low confidence, by the digital model, e.g., data generated by the system in new, previously unseen, operating conditions are selected and used to update the model. This approach avoids the costly transmission of data with low additional information content. In addition to novel data, the most anomalous data, as determined by the prediction error of the DT, are collected during the operation of the physical asset. The data are automatically labelled by comparison to the fleet data set and made available for expert evaluation to aid root-cause analysis. Anomalous data are removed from the update data set to avoid learning emerging faults. 

The main contributions of the proposed approach are: (1) a light-weight DT capable of running online in constrained systems, (2) a DT-based data prioritisation to collect the most relevant data, and (3) a robust anomaly detection for individual systems that takes advantage of the fleet information.

Two key features of the proposed approach are the selection of appropriate data to update the digital model and the detection of anomalous data. These topics are discussed in the remainder of this section.

\subsection{Data selection}
The DT should be routinely updated with data generated by the physical system in order to keep the twin up-to-date and be able to identify unusual behaviours. For resource-limited systems, such an update cannot be performed online; the data must be stored or transmitted to update the DT on a more powerful computing system. Several factors can limit the amount of data collected by resource-limited systems, e.g., data transmission cost, limited bandwidth, limited storage or energy capacity. This problem requires the development of methods to select high-quality data that are information-rich for the desired task.



The data selection problem is present in the field of active learning. Active learning considers the problem of selecting unlabelled data to be labelled for the purpose of model training. The main challenge is how to select the most informative data, such that the performance of a model is maximised. Research on this problem has resulted in different strategies to select data such as models' disagreement \cite{burbidge2007active}, expected model change maximization \cite{cai2013maximizing} and uncertainty sampling \cite{yang2015multi}. In uncertainty sampling, samples whose class assignment or prediction are the most uncertain are selected. Due to the high memory and computational requirements of the first two approaches, these are not suitable for resource-limited systems. Therefore, the proposed solution in this paper selects segments of data based on the model uncertainty to update the DT.


\subsection{Anomaly detection}
An anomaly, also called an outlier, is a pattern in the data that deviates from a defined notion of normality. 
At a high level, the problem of anomaly detection consists of learning a region or representation of normal behaviour and identifying data that does not belong to that region. There are several challenges that make this task difficult \cite{chandola2009anomaly}. First, a clear distinction between anomalies and normal behaviour is not well defined in many situations. 
Second, depending on the approach used to define normality, a large amount of labelled data, which can be difficult and expensive to collect, is required. Finally, when the system under analysis changes over time, e.g., concept drift, previous definitions of normality might not be adequate to identify anomalies. Depending on the application or domain, additional challenges may be presented. 


Multiple solutions have been proposed to solve the problem of anomaly detection \cite{pimentel2014review}. A considerable amount of available anomaly detection methods assume that the data are stored and significant computational resources are available. However, for many real-world systems such as streaming data systems, where data grows infinitely, or for resource-constrained systems, such methods cannot be executed \cite{rajasegarar2008anomaly}. 
Such limitations have motivated the development of new methods to detect anomalies within resource-constrained systems. Efficient on-device anomaly detectors based on model comparison \cite{attia2015device} and remote processing \cite{amontamavut2012separated} have been proposed. 

Anomaly detection has also been studied in groups of systems, e.g., a fleet of vehicles. In this scenario, although the systems all have similar behaviour, each system is unique due to variations in manufacturing, usage and degradation. Learning a single model of normality with the average fleet behaviour can result in a poor overall detection performance due to the variability of individual systems. An alternative is to learn a representation of normality for each system in the group. Although anomalies can be detected with such an approach, considering data from all the systems in the group helps to mitigate the problem of unusual data and data availability. For instance, unknown operating conditions for a particular system might be seen as normal when data from other systems are considered.


The solution proposed in this paper uses a combination of local and centralised anomaly detection in order to exploit the benefits of fleet data while minimising handling data costs. At the system level, the digital model is used to identify potentially anomalous segments of data in each system. At the group level, a centralised system with access to historical data from the group is used to refine the anomaly detection performed by each system. In contrast to other available methods, the approach proposed here allows the update of individual models and the identification of anomalies without collecting or transmitting all data, and hence facilitates a practical implementation. 

The rest of the paper is organised as follows. Section \ref{sec:ProbDefinition} presents the problem solved in this paper and introduces two case studies. Section \ref{sec:Solution} explains in detail the proposed solution. Finally, results and conclusions are given in sections \ref{sec:Results} and \ref{sec:Conclusion}, respectively.

\section{Problem Definition}\label{sec:ProbDefinition}

In this paper, we focus on the development of a data-driven DT for groups of resource-limited systems. Specifically, we consider systems with the following characteristics. The system is not capable of executing tasks with a high computational load such as running full-physics models, updating digital models or executing complex anomaly detection methods. The system cannot store or transmit all the data to be processed by a system with more resource availability. Moreover, data cannot be continuously transmitted to another system.

We consider complex non-linear and dynamic systems subject to slow nominal intrinsic degradation, rare but acceptable extrinsic disturbances and the risk of abnormal intrinsic anomalies / faults. Given these characteristics, the challenge is to develop a framework to compute and update a DT that is capable of accurately simulating a varying physical system throughout its lifetime. To maintain a digital representation of the system, the problems of selecting, managing and analysing collected data must be addressed.

\subsection{Case studies}

\begin{figure}[ht]
\includegraphics[width=0.49\textwidth]{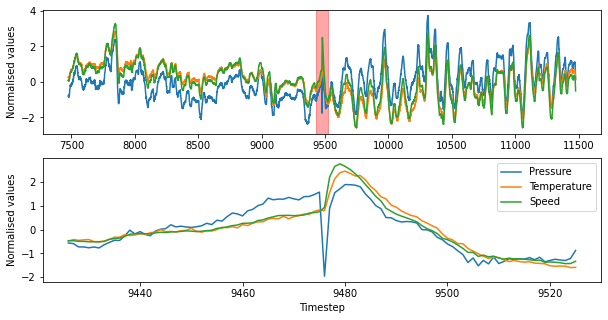}
\caption{\textbf{Case study 1, observed anomaly.} Top: normalised multivariate time series data recorded during flight. Bottom: anomaly observed in a pressure signal (highlighted in top figure).}
\label{fig:P30Anomaly}
\end{figure}

The solution presented in this paper is demonstrated with two real data case studies, both using data from aerospace gas turbine engines. The two case studies are presented below:

\textbf{Anomaly detection}. The first case study considers an in-service gas turbine engine with a known fault. Data recorded during 116 consecutive flights are used to train the DT and test its ability to perform anomaly detection. The dataset consists of 1Hz time series data with 12 channels - each channel records data from a different sensor placed on the engine. Fault symptoms are observed in one flight, see Fig. \ref{fig:P30Anomaly}. This figure shows a permanent change in the relationship between signals at the moment the fault occurs. The detection of this fault is used to illustrate the capacity of the proposed solution to identify anomalous data. 

\begin{figure}[ht]
\includegraphics[width=0.49\textwidth]{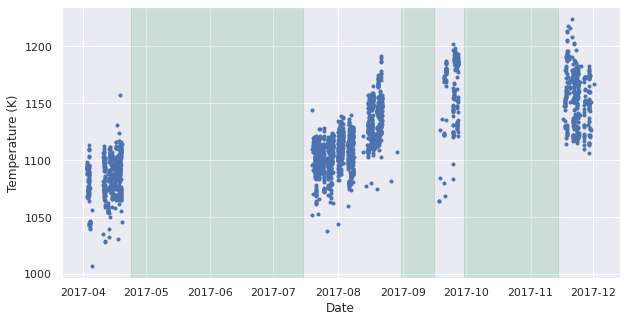}
\caption{\textbf{Case study 2, observed degradation and maintenance.} Mean engine temperature recorded at a fixed shaft speed. As the engine degrades  over time it becomes less efficient and more energy is required to achieve a demanded shaft speed. As a result, for a given shaft speed, a higher engine temperature is recorded. Highlighted regions show periods where a major maintenance action was performed.}
\label{fig:ETOPSTestsDate}
\end{figure}

\textbf{Digital Twin update}. The second case study demonstrates the ability of the proposed solution to update the DT to accurately simulate a physical system when its behaviour changes through time. Data were recorded from an aerospace gas turbine engine over multiple runs on a testbed over a period of several months. During this period the engine showed signs of nominal degradation and several maintenance actions were performed, see Fig. \ref{fig:ETOPSTestsDate}. The degradation and maintenance caused the dynamic behaviour of the system to change continuously over time. As in the previous case study, the data consist of 1Hz time series data, but here over 80 data channels are available. 

\section{Solution}\label{sec:Solution}

Digital Twins should represent the real system throughout its life cycle. In general, a model prediction can deviate from the observed data due to the following reasons: (1) degradation: the system behaviour changes due to expected degradation, (2) unseen operating conditions: the system operates under conditions not previously observed, and (3) system fault: unexpected behaviour is observed caused by a system fault. To update the DT, data must be constantly collected from the physical system. Moreover, to stop a DT from tracking (learning) emerging faults, the data used to update the DT must be analysed. This is a challenge faced in on-line learning where a model is constantly updated from a data stream. Here, a model must learn changes in data distribution but avoid learning anomalous behaviours. Solutions based on changes in the data distribution have been proposed \cite{saurav2018online}. In these approaches, when a model prediction deviates from the observed data over a significant period of time, it is assumed that the data distribution has changed. However, as presented in Section \ref{sec:Results}, depending on the problem, a permanent change of the system behaviour can be considered as an anomaly. 

\begin{figure}[ht]

\includegraphics[width=0.49\textwidth]{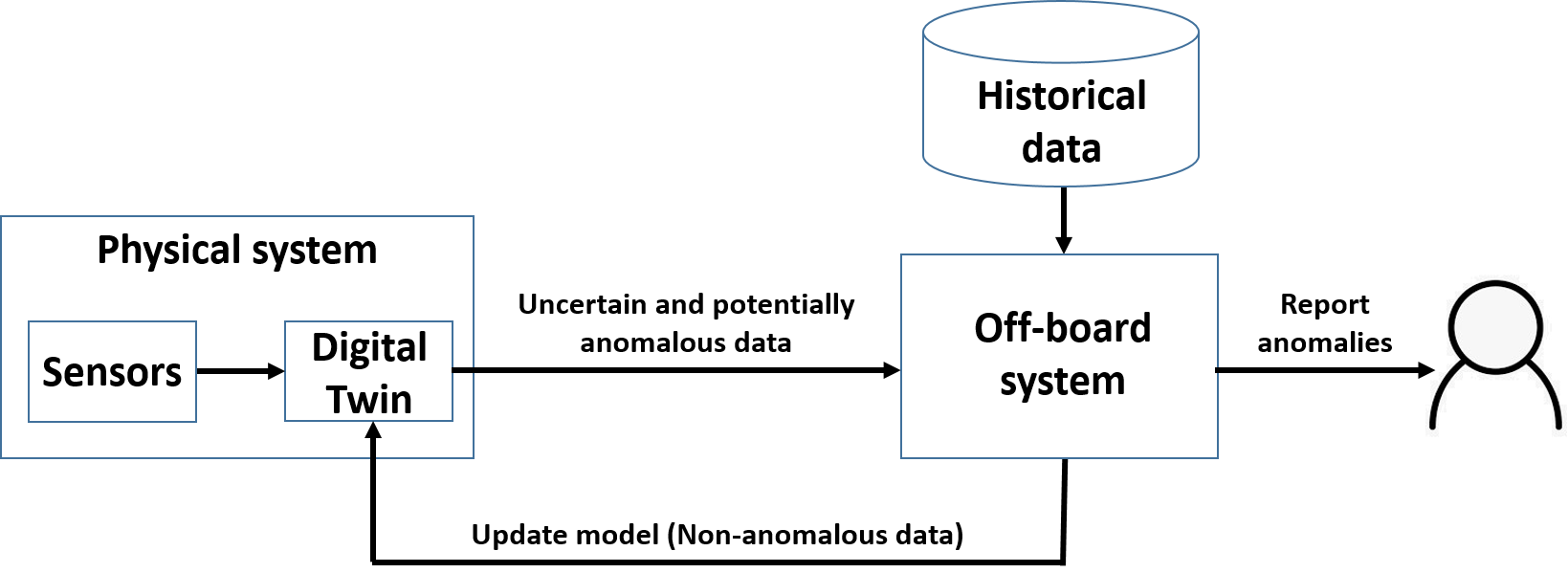}
\caption{\textbf{Overview of proposed solution.} Uncertain and potentially anomalous data are selected on-board the physical system by using the Digital Twin. Data are sent to the off-board system for analysis, with access to the group's historical data. The off-board system identifies and presents the anomalous data to a user for further analyses. Normal (non-anomalous) data are used to update the Digital Twin.}
\label{fig:Flowchart}
\end{figure}

We propose a closed-loop approach: selected data collected from the physical system are sent to an anomaly detector, which also has access to group data. The selection is based on the model's uncertainty and data anomaly. This allows us to collect data that are not well understood by the model, e.g., data from new operating conditions. The non-anomalous uncertain data are then used to update the DT. By improving the capability of the DT to predict behaviours in different operation conditions, the DT increases its capability to identify anomalous behaviours with greater accuracy as explained in the next section. In addition, access to the group's data allows us to identify behaviours caused by degradation or previously unseen operating conditions, for a particular system, that could be flagged as anomalous even if such behaviour is normal at the group level. The combination of the DT update with selected data and the access to the group's data increases the robustness of the anomaly detection. An overview of the proposed approach is shown in Fig. \ref{fig:Flowchart}.

Since we consider resource-constrained systems, collecting or transmitting all the data is not feasible. We use a light-weight DT capable of running on-board the physical system. The DT has the ability to identify and collect both potentially anomalous data, and data that are not well understood, i.e., data associated with a high prediction uncertainty. An example of the latter includes data at previously unseen operating conditions. In this paper, we refer to the combination of on-board DT and software used for data collection as the on-board system. Data collected by the on-board system are then analysed by a centralised system, referred to as off-board system, with access to greater computational resources and the group's historical data. The off-board system identifies whether the data returned by the on-board system are anomalous at the group level. This data can then be presented to experts for assessment of label accuracy, root-cause analysis and sanctioning of alerts. Data not identified as anomalous are used to update the DT. 
In the remainder of this section, the on-board and off-board systems are presented.

\subsection{On-board system}

Here, we present the approaches used to model the physical system and for data selection. Due to the system's limited processing and storage capacity, a custom-designed deep neural network was selected to model the system's behaviour. In contrast to other computationally expensive models, e.g., full-physics models, deep neural networks can be efficiently implemented to run on systems with limited resources using techniques such as quantisation and pruning \cite{reddy2017real}. The key elements of the network that allow the selection of uncertain and potentially anomalous data are discussed below. A bespoke Convolution Neural Network (CNN) has been designed to capture the dynamic behaviour of multi-input and non-linear physical systems. CNNs are attractive in this application due to ability to compress the models for efficient execution on embedded systems \cite{sze2017efficient}. The model is trained to represent the physical system behaviour with a parsimonious set of features, aiding fast run-time execution.

\begin{figure}[ht]
\centering
\includegraphics[width=0.3\textwidth]{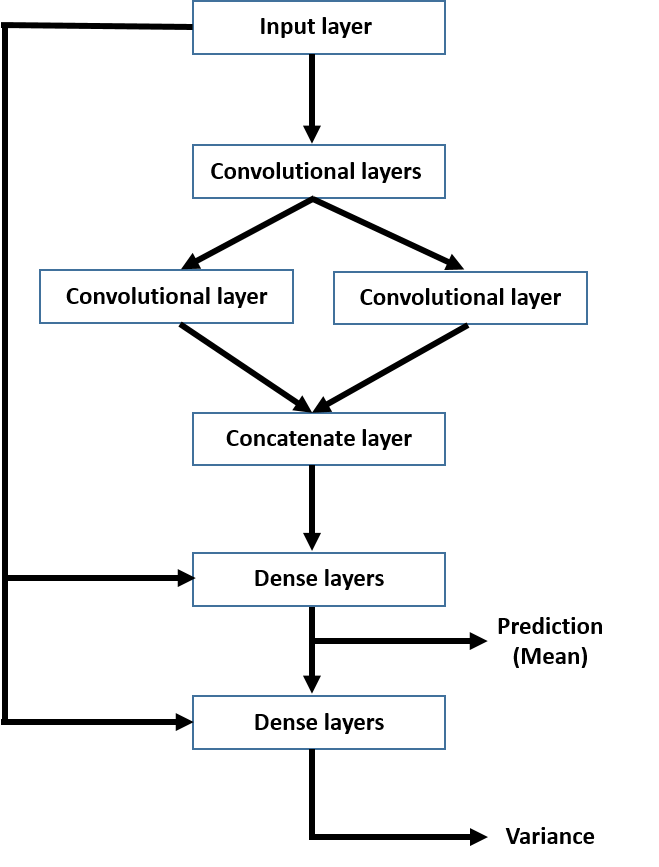}
\caption{\textbf{Overview of the network architecture.} Convolutional layers are used to extract features from individual signals. To improve performance at minimal computational cost, the network takes advantage of skip connections.}
\label{fig:NN_illustration}
\end{figure}

At each sample time, $k$, the deep neural network receives a window of time-series data, $X$, (signals from the system's sensors) as an input and makes a prediction on the distribution of the output signal, $y$, which is assumed to be Gaussian;
\begin{equation}
	y(k) \sim \mathcal{N} \left( y(k) |  f(X,  \boldsymbol{\theta}), \sigma(X,  \boldsymbol{\theta}) \right)
\end{equation}
where $\boldsymbol{\theta}$ are the network weights, 
\begin{eqnarray*}
	X &=& \left[\mathbf{x}_1, \mathbf{x}_2, \ldots, \mathbf{x}_m \right] \in \mathcal{R}^{N \times m},\\ 
	\mathbf{x}_i  &=& \left[x_i(k-N), x_i(k-N+1), \ldots, x_i(k-1) \right]^T \in \mathcal{R}^{N \times 1}
\end{eqnarray*}
is a vector containing the previous $N$ data points of the $i$'th input data stream and $m$ is the number of channels in the input data. 

The Gaussian distribution is characterised by its mean, $\mu = f(X,  \boldsymbol{\theta})$, and variance  $\sigma = \sigma(X, \boldsymbol{\theta})$, which are dependant on the input data, $X$, and are estimated simultaneously by the deep neural network, see Fig. \ref{fig:NN_illustration}. The model weights $\boldsymbol{\theta}$ are trained via minimisation of the negative log likelihood given by
\begin{equation} \label{eqn: loss}
	\mathcal{L} = - \ln \left( \prod_{n=1}^B \mathcal{N} \left( y(n) |  \mu(n), \sigma(n) \right) \right), 
\end{equation}
using a stochastic gradient descent based algorithm which is fed batches of data at each iteration with batch size $B$. The details of the neural network architecture, training and implementation are omitted in this paper and the interested reader is referred to \cite{hartwell2021inflight}.





The purpose of computing the variance is twofold: (1) identify data that are uncertain to the model, and (2) compute a standardised Euclidean distance to identify anomalous data. Large prediction variance (low confidence) can be attributed to input data that is distributed differently to input data given at training time. To improve the prediction in such regions of the input space, and hence improve the robustness of the DT, data in a window around the most uncertain predictions are collected for analysis by the off-board system. If the returned segments of data are not anomalous with respect the fleet data they are used to update the DT as explained in Section \ref{subsec:DTUpdate}. This update is critical to avoid the masking of anomalous behaviours by previously unseen normal behaviours. 

To identify possible anomalies, the mean and variance are used to calculate the standardised Euclidean distance:

\begin{equation}\label{eq:Mahalanobis}
d(y(k), \mu(k)) = \sqrt{\frac{(y(k)- \mu(k))^2}{\sigma^2(k)}}.
\end{equation}

Intuitively, the standardised Euclidean distance penalises prediction errors that are large relative to the predicted standard deviation. This avoids labelling nominal data in unknown conditions as faults. Similar to the uncertain data, windows centred at large standardised Euclidean distance are collected. Since the physical system memory is limited, the top $N$ most uncertain and anomalous windows are retained during operation. These windows are then sent to the off-board system to be analysed when a data link is available, avoiding the need for a decision on anomalous behaviour to be made on-board. Note that $N$ is chosen to store the maximum amount of data. Hence, it is determined by the systems data storage constraints. 

\subsection{Off-board system}

The objective of the off-board system is to collect data returned by on-board system to update the digital model and identify anomalies.
Two key elements differentiate the on-board and off-board systems: (1) access to more computational resources, and (2) access to the individual asset and group's historical data. More computational resources mean that the off-board system can run more computationally demanding anomaly detection routines. Moreover, running time is not crucial in the off-board system in contrast to the on-board system, where data are constantly received and have to be analysed online. Robustness in the anomaly detection is achieved by accessing the group's historical data to allow the off-board system to identify data that are anomalous to a particular system but not to the group.

The selection of the method used off-board is highly problem dependant. 
Therefore, we only give an overview of the method used for the case studies.

As discussed in Section \ref{sec:ProbDefinition}, the data analysed in the case studies are multivariate time series. Features were extracted from the time series of all historical data. Specifically, a combination of features, obtained by using Kernel PCA \cite{yang2005pca} that reflects the correlation between signals and statistical features from individual time series were used. These statistical features can be designed and updated based on knowledge from anomalous data collected by the off-board system over time. 
The extraction of these features is computationally expensive, hence such a method cannot be run on-board. The features are used to train a one-class Support Vector Machine (SVM). The SVM can be trained in such a way that rare nominal events not learnt by the on-board system, and hence flagged as possible anomalies, are classified as non-anomalous data. 

\subsection{Digital Twin update}\label{subsec:DTUpdate}

\begin{figure}[ht]
\includegraphics[width=0.49\textwidth]{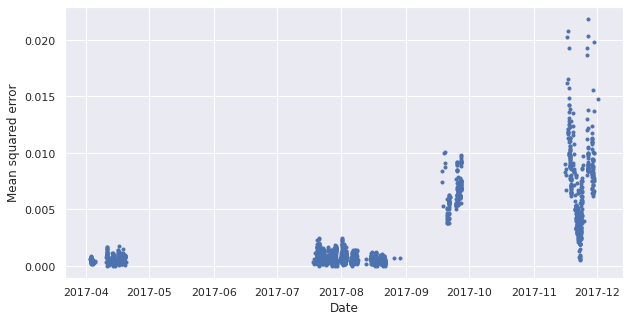}
\caption{Increase of prediction error due to system overhaul and degradation.}
\label{fig:degradation}
\end{figure}

When the physical system changes its behaviour due to factors such as degradation, the DT should be updated to describe the new behaviour. To show the effects of a change in the system behaviour, consider the example shown in Fig. \ref{fig:degradation}. The example shows the mean squared prediction error of a predicted signal over several runs from the same engine during a time period of multiple months. The predictions are made by a model trained with data from engine runs performed before the initial date shown in the figure. As noted in Section \ref{sec:ProbDefinition}, the engine undergoes maintenance during each large period of inoperation. As a result, the dynamics of the engine change at each maintenance event such that the current model is not able to predict the engine behaviour.

The example above shows the necessity of updating the DT with the most recently collected data. In other words, the deep neural network has to be updated with a relatively small dataset compared to the dataset initially used to train the model. This problem has been addressed in transfer learning by using fine-tuning \cite{yosinski2014transferable}. Two major problems of using a small dataset to update a pre-trained model are overfitting and catastrophic forgetting \cite{mccloskey1989catastrophic}, i.e., a model can forget previously learnt knowledge when is trained with new information. Parameter regularisation methods have been used to mitigate this problem. By restricting the ability of the network to learn, the problem of overfitting can be reduced. Formally, these methods minimise a loss function of the form:

\begin{equation}
\tilde{\mathcal{L}} = \mathcal{L} + \sum_j \Omega_j \|\theta_j - \theta^*_j\|_2^2,
\end{equation}

\noindent where $\mathcal{L}$ is the original loss function, here given by (\ref{eqn: loss}), $\theta_j$ is the $j$'th network weight, $\theta_j^*$ is the $j$'th weight of the pre-trained network and $\Omega$ is a hyper-parameter that controls the regularisation strength. One of the most common types of regularisation is $L^2$ regularisation, where all the parameters are forced towards zero, i.e., $\theta^*_j = 0 \; \forall \;j$. Other approaches maintain the values of the original network, i.e., $\theta^*$, and compute the regularisation strength of individual weights based on how important they are to the previous knowledge \cite{kirkpatrick2017overcoming}. 

Two different regularisers were considered to update the model when new data are received from the on-board system: $L^2$ and $L^2$-$SP$ \cite{daume2009frustratingly}. Results obtained from several tasks show that these regularisers are competitive compared to more complex and computationally demanding approaches \cite{li2018explicit}. The $L^2$-$SP$ regulariser penalises all the parameters with the same factor, i.e., $\Omega_j = \alpha \; \forall \;j$ and $\alpha \in \mathbb{R}$. All the weights are hence forced to remain close to those of the pre-trained network. In contrast to a model updated with the $L^2$ regulariser, a model updated with the $L^2$-$SP$ regulariser is expected to remember previously seen behaviours while learning new information.  

\section{Results}\label{sec:Results}

In this section, we demonstrate the ability of the proposed solution to model a physical system, identify anomalous data, and update the digital model.

\subsection{Anomaly detection} \label{subsec:DT}

The ability to detect anomalous data is demonstrated with the first case study presented in Section \ref{sec:ProbDefinition}. The dataset was divided into training and testing sets. The data known to contain fault symptoms was placed in the testing set. After training the neural network, the test dataset was used to test the ability of the network to predict the behaviour of the engine and to identify anomalies. The on-board system was run on an ARM Cortex-A7 micro-processor (as used in a production monitoring system). The time series in the testing set were fed to the neural network to predict the value of the target signal. The standardised Euclidean distance or score was computed at each timestep to identify the most anomalous data. 

\begin{figure}[ht]
\includegraphics[width=0.49\textwidth]{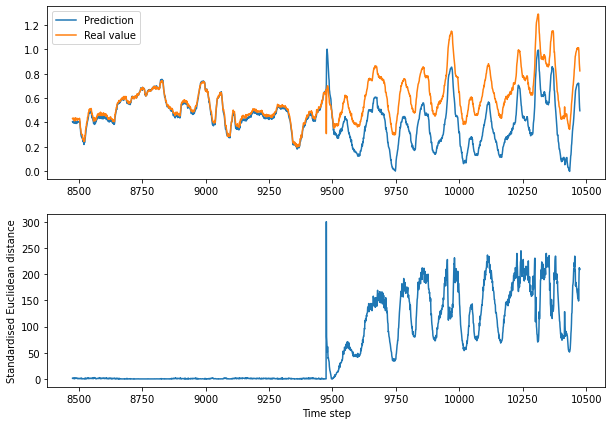}
\caption{Top: The real and predicted values of the target signal (pressure). A fault occurs at time step 9476 causing a change in the system behaviour. Bottom: Standardised Euclidean distance. The standardised Euclidean distance is used to select the most anomalous data.}
\label{fig:P30Mahalanobis}
\end{figure}

Fig. \ref{fig:P30Mahalanobis} shows the prediction and score of a segment of the dataset containing the known fault. It is possible to see that once the fault occurs, the behaviour of the engine changes and therefore the model is not capable of predicting the new behaviour. This results in a high score after the fault. 

From each time series in the testing set, the most anomalous and uncertain windows were collected. The amount of data returned by the on-board system is limited by the transmission bandwidth. To maximise the amount of data collected, a threshold is not used by the on-board system to select anomalous data. Instead, the on-board system uses a ranking system where the most anomalous data, based on the standardised Euclidean distance, is prioritised. In the case study, the top 100 anomalous and uncertain windows were collected for each flight. The data were then analysed by the off-board system to remove non-anomalous data. Due to limited data availability, the off-board system anomaly detector was trained with historical data, collected over a long period of time from one engine. Such historical data can contain rare behaviours that the current DT may not have adequately learned. Hence, the off-board system is capable of identifying segments of nominal data flagged as anomalous by the on-board system. Note that although just one engine was considered, the proposed approach is designed to use data from a group of systems or fleet. From 1400 analysed windows, collected by the on-board system over the preceding 14 flights, the off-board system identified 10 windows as anomalies, including the known anomaly. This demonstrates the ability of the proposed solution to reduce the workload of the user. A further analysis was done in the rest of the windows detected as anomalous. This analysis consists of comparing the data recorded from one engine to the data recorded from its sister engine (the other engine used during flight). A high residual between the data recorded from both engines indicates an anomalous behaviour in one of the engines. Of the 10 flagged windows, 5 windows (including the known anomaly) presented high residual between engines. This confirms the known anomalous behaviour and possible previously unknown anomalies.   

\begin{figure}[ht]
\includegraphics[width=0.49\textwidth]{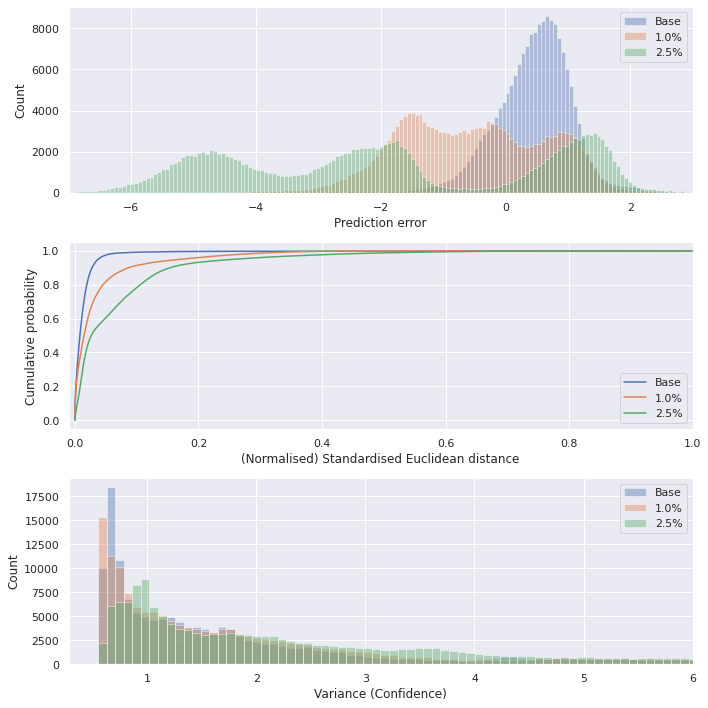}
\caption{Effects of different anomalies with different magnitudes (as a percentage of input range) in the deep neural network inputs. Top: Histogram of prediction errors. Centre: Cumulative distribution of normalised standardised Euclidean distance. Bottom: Histogram of the predicted variance. As the noise levels in the input space increases the confidence of the network decreases, i.e. there is an increment in the variance.}
\label{fig:NoiseInput}
\end{figure}

The example above shows the detection of real anomalies in the predicted signal. To illustrate the detection of anomalies in the inputs of the on-board neural network, anomalies with different sizes were synthetically injected into the input data, see Fig. \ref{fig:NoiseInput}. The top figure shows a distribution of the prediction error, i.e., the difference between predicted and real value, as the size of the anomalies increases. As expected, an increment of the anomaly size results in an increment of the prediction error. The figure in the centre shows the cumulative distribution of the standardised Euclidean distance. This figure demonstrates that for a given communication capacity, the average error corresponding to the returned anomalies will increase.

Finally, the last plot shows the effects of unknown inputs in the confidence of the prediction. The confidence of the prediction decreases as the inputs become more anomalous. This confidence reduction penalises prediction errors for previously unknown inputs. However, as presented in the results of the first case study, such a penalty does not avoid detecting anomalies for sufficiently large anomalous inputs.

\subsection{Digital Twin update}

Here we present the results of updating the digital model with data collected from the physical system. This corresponds to the second case study presented in Section \ref{sec:ProbDefinition}. Moreover, we compare the regularisation approaches presented in Section \ref{subsec:DTUpdate} in terms of prediction accuracy and the capacity to remember previous behaviours. 

\begin{figure}[ht]
\includegraphics[width=0.49\textwidth]{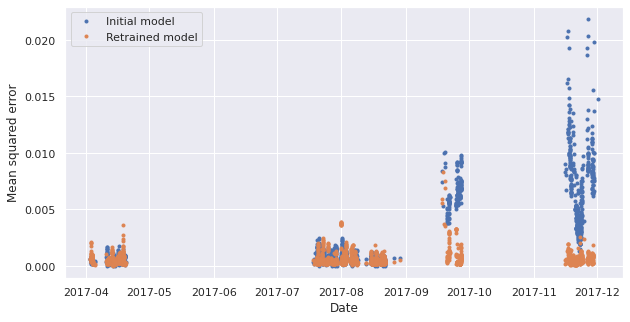}
\caption{Mean squared error of predictions made by the initial and iteratively retrained model.}
\label{fig:ModelsPredComparison}
\end{figure}

An iterative update strategy is used to update the DT. An initial model is trained with all data from runs performed before April 17, and is used to select the unusual data from proceeding 10 runs (after the training data). The collected data were used to update the DT. Once the model was updated, data from the next 10 runs were processed and collected to again update the DT. This procedure was repeated until all runs were processed. The prediction errors of the initial model and the updated model are shown in Fig. \ref{fig:ModelsPredComparison}. The significant increase in error observed after 2017-09 for the initial model is due to a maintenance action on the engine. The results show that by updating the DT the behaviour of the physical system can be predicted, despite changes in the system. 

To illustrate how the incorrect simulation of the physical system can affect the identification of anomalies, synthetic anomalies (spikes of fixed length and amplitude based on the signal's local standard deviation) were injected in randomly selected runs. The runs were divided into two groups: before and after the September maintenance event. As shown in Fig. \ref{fig:ModelsPredComparison}, when the DT is not updated, the behaviour of the system is not predicted accurately. Therefore, poor performance in the detection of anomalies is expected, and this is shown in the results. When the DT is not updated, 22 of the synthetic anomalies are detected before the maintenance event and only one after. In contrast, when it is updated, 28 and 23 anomalies are detected before and after the overhaul, respectively. The results show an improvement in the number of anomalies detected when the DT is routinely updated. As expected, a more noticeable improvement is made on the data after the maintenance event.\\

\begin{figure}[ht]
\includegraphics[scale=0.31]{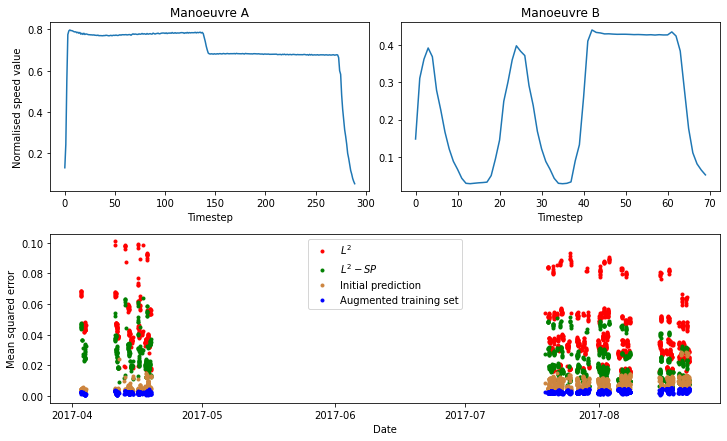}
\caption{Top: manoeuvres representing different behaviours of the engine. Bottom: Prediction of manoeuvre A after retraining a model with manoeuvre B using different training approaches.}
\label{fig:Manoeuvres}
\end{figure}

The results presented above show that when a model is routinely updated, accurate prediction of engine behaviour is achieved. However, they do not show how the model is affected in terms of forgetting previous behaviours. To see this effect, two different manoeuvres or operating profiles, labelled A and B, were extracted from each engine run, see Fig. \ref{fig:Manoeuvres} (top). A model was initially trained with manoeuvre A, and it is assumed that no manoeuvre B data are available. Then, the model was iteratively retrained with data from manoeuvre B following the procedure presented above. Three different approaches were used to retrain the model: the $L^2$ regularisation, the $L^2-SP$ regularisation, and an augmented training data approach which includes historic data from manoeuvre A. Fig. \ref{fig:Manoeuvres} (bottom) shows the prediction of manoeuvre A after retraining the model. The prediction of manoeuvre B did not vary significantly between approaches and is not shown.

As expected, the results show that when the model is updated with the $L^2$ regulariser, the model forgets how to predict manoeuvre A. The forgetting is significantly reduced when using the $L^2-SP$ regulariser. Adding data from manoeuvre A during the retraining improves the capacity of the model to remember manoeuvre A. While this approach does not represent a challenge in the toy example presented above, it raises new challenges when data must be selected from a large pool of historical engine behaviours which might not represent the current system behaviour. 

\section{Conclusion}\label{sec:Conclusion}

Digital Twins are used in a wide range of areas for management, optimisation and decision support of physical systems. To accurately represent its physical counterpart, DTs must be routinely updated with data collected from the physical system to be able to identify unusual behaviours in the physical system. This can be a significant challenge for systems with constrained resources. Moreover, not all the collected data are suitable to update the DT, e.g., anomalies related to emerging faults must be handled carefully, particularly in our chosen application of system health monitoring. This paper presents a novel 2-stage framework to keep a data-driven DT synchronised with its the physical system counterpart. The proposed solution allows real-time asset monitoring, and the selection of high-quality data for remote update of a DT using fault-free fleet data. Results from a gas turbine engine case study show the capacity of the solution to accurately simulate the behaviour of an engine throughout its life. To further reduce the computational and data requirements for updating, interesting future research directions include the definition of (optimal) criteria on when to perform model updates and the introduction of physics-based constraints to the machine learning update process.



\bibliographystyle{unsrt}
\bibliography{references} 

\begin{thebibliography}{10}

\bibitem{tao2018digital}
F.~Tao, H.~Zhang, A.~Liu, and A.~Y.C. Nee.
\newblock Digital twin in industry: State-of-the-art.
\newblock {\em IEEE Transactions on Industrial Informatics}, 15(4):2405--2415,
  2018.

\bibitem{glaessgen2012digital}
E.~Glaessgen and D.~Stargel.
\newblock The digital twin paradigm for future nasa and u.s. air force
  vehicles.
\newblock In {\em Proc. of AIAA/ASME/ASCE/AHS/ASC structures, structurual
  dynamics and materials conference}, page 118, 2012.

\bibitem{barricelli2019survey}
B.~R. Barricelli, E.~Casiraghi, and D.~Fogli.
\newblock A survey on digital twin: Definitions, characteristics, applications,
  and design implications.
\newblock {\em IEEE access}, 7:167653--167671, 2019.

\bibitem{leng2021digital}
J.~Leng, D.~Wang, W.~Shen, X.~Li, Q.~Liu, and X.~Chen.
\newblock Digital twins-based smart manufacturing system design in industry
  4.0: A review.
\newblock {\em Journal of manufacturing systems}, 60:119--137, 2021.

\bibitem{DTHub}
The {DTH}ub.
\newblock http://https://digitaltwinhub.co.uk/.
\newblock Accessed: 2021-06-15.

\bibitem{qi2018digital}
Q.~Qi and F.~Tao.
\newblock Digital twin and big data towards smart manufacturing and industry
  4.0: 360 degree comparison.
\newblock {\em Ieee Access}, 6:3585--3593, 2018.

\bibitem{tuegel2011reengineering}
E.~J. Tuegel, A.~R. Ingraffea, T.~G. Eason, and S.~M. Spottswood.
\newblock Reengineering aircraft structural life prediction using a digital
  twin.
\newblock {\em International Journal of Aerospace Engineering}, 2011, 2011.

\bibitem{magargle2017simulation}
R.~Magargle, L.~Johnson, P.~Mandloi, P.~Davoudabadi, O.~Kesarkar,
  S.~Krishnaswamy, J.~Batteh, and A.~Pitchaikani.
\newblock A simulation-based digital twin for model-driven health monitoring
  and predictive maintenance of an automotive braking system.
\newblock In {\em Proc. of the International Modelica Conference}, pages
  35--46. Link{\"o}ping University Electronic Press, 2017.

\bibitem{li2017dynamic}
C.~Li, S.~Mahadevan, Y.~Ling, S.~Choze, and L.~Wang.
\newblock Dynamic bayesian network for aircraft wing health monitoring digital
  twin.
\newblock {\em Aiaa Journal}, 55(3):930--941, 2017.

\bibitem{wang2019digital}
J.~Wang, L.~Ye, R.~X. Gao, C.~Li, and L.~Zhang.
\newblock Digital twin for rotating machinery fault diagnosis in smart
  manufacturing.
\newblock {\em International Journal of Production Research},
  57(12):3920--3934, 2019.

\bibitem{chakraborty2021machine}
S.~Chakraborty and S.~Adhikari.
\newblock Machine learning based digital twin for dynamical systems with
  multiple time-scales.
\newblock {\em Computers \& Structures}, 243:106410, 2021.

\bibitem{fink2020potential}
O.~Fink, Q.~Wang, M.~Svensen, P.~Dersin, W.~Lee, and M.~Ducoffe.
\newblock Potential, challenges and future directions for deep learning in
  prognostics and health management applications.
\newblock {\em Engineering Applications of Artificial Intelligence}, 92:103678,
  2020.

\bibitem{burbidge2007active}
R.~Burbidge, J.~J. Rowland, and R.~D. King.
\newblock Active learning for regression based on query by committee.
\newblock In {\em Proc. of IDEAL}, pages 209--218. Springer, 2007.

\bibitem{cai2013maximizing}
W.~Cai, Y.~Zhang, and J.~Zhou.
\newblock Maximizing expected model change for active learning in regression.
\newblock In {\em Proc. of ICDM}, pages 51--60. IEEE, 2013.

\bibitem{yang2015multi}
Y.~Yang, Z.~Ma, F.~Nie, X.~Chang, and A.~G. Hauptmann.
\newblock Multi-class active learning by uncertainty sampling with diversity
  maximization.
\newblock {\em International Journal of Computer Vision}, 113(2):113--127,
  2015.

\bibitem{chandola2009anomaly}
V.~Chandola, A.~Banerjee, and V.~Kumar.
\newblock Anomaly detection: A survey.
\newblock {\em ACM computing surveys (CSUR)}, 41(3):1--58, 2009.

\bibitem{pimentel2014review}
M.~A.F. Pimentel, D.~A. Clifton, L.~Clifton, and L.~Tarassenko.
\newblock A review of novelty detection.
\newblock {\em Signal Processing}, 99:215--249, 2014.

\bibitem{rajasegarar2008anomaly}
S.~Rajasegarar, C.~Leckie, and M.~Palaniswami.
\newblock Anomaly detection in wireless sensor networks.
\newblock {\em IEEE Wireless Communications}, 15(4):34--40, 2008.

\bibitem{attia2015device}
M.~B. Attia, C.~Talhi, A.~Hamou-Lhadj, B.~Khosravifar, V.~Turpaud, and
  M.~Couture.
\newblock On-device anomaly detection for resource-limited systems.
\newblock In {\em Proc. of ACM/SIGAPP}, pages 548--554, 2015.

\bibitem{amontamavut2012separated}
P.~Amontamavut, Y.~Nakagawa, and E.~Hayakawa.
\newblock Separated linux process logging mechanism for embedded systems.
\newblock In {\em Proc. of RTCSA}, pages 411--414. IEEE, 2012.

\bibitem{saurav2018online}
S.~Saurav, P.~Malhotra, V.~TV, N.~Gugulothu, L.~Vig, P.~Agarwal, and G.~Shroff.
\newblock Online anomaly detection with concept drift adaptation using
  recurrent neural networks.
\newblock In {\em Proc. of the ACM India Joint International Conference on Data
  Science and Management of Data}, pages 78--87, 2018.

\bibitem{reddy2017real}
B.~Reddy, Y.~Kim, S.~Yun, C.~Seo, and J.~Jang.
\newblock Real-time driver drowsiness detection for embedded system using model
  compression of deep neural networks.
\newblock In {\em Proc. of CVPR}, pages 121--128, 2017.

\bibitem{sze2017efficient}
V.~Sze, Y.~Chen, T.~Yang, and J.~S. Emer.
\newblock Efficient processing of deep neural networks: A tutorial and survey.
\newblock {\em Proceedings of the IEEE}, 105(12):2295--2329, 2017.

\bibitem{hartwell2021inflight}
Adam Hartwell, Felipe Montana, Will Jacobs, Visakan Kadirkamanathan, Andrew~R
  Mills, and Tom Clark.
\newblock In-flight novelty detection with convolutional neural networks.
\newblock {\em arXiv preprint arXiv:2112.03765}, 2021.

\bibitem{yang2005pca}
K.~Yang and C.~Shahabi.
\newblock A pca-based kernel for kernel pca on multivariate time series.
\newblock In {\em Proc. of ICDM}, 2005.

\bibitem{yosinski2014transferable}
J.~Yosinski, J.~Clune, Y.~Bengio, and H.~Lipson.
\newblock How transferable are features in deep neural networks?
\newblock In {\em Advances in neural information processing systems}, pages
  3320--3328, 2014.

\bibitem{mccloskey1989catastrophic}
M.~McCloskey and N.~J. Cohen.
\newblock Catastrophic interference in connectionist networks: The sequential
  learning problem.
\newblock In {\em Psychology of learning and motivation}, volume~24, pages
  109--165. Elsevier, 1989.

\bibitem{kirkpatrick2017overcoming}
J.~Kirkpatrick, R.~Pascanu, N.~Rabinowitz, J.~Veness, G.~Desjardins, A.~A.
  Rusu, K.~Milan, J.~Quan, T.~Ramalho, A.~Grabska-Barwinska, et~al.
\newblock Overcoming catastrophic forgetting in neural networks.
\newblock {\em Proc. of the national academy of sciences}, 114(13):3521--3526,
  2017.

\bibitem{daume2009frustratingly}
H.~Daum{\'e}~III.
\newblock Frustratingly easy domain adaptation.
\newblock {\em arXiv preprint arXiv:0907.1815}, 2009.

\bibitem{li2018explicit}
X.~Li, Y.~Grandvalet, and F.~Davoine.
\newblock Explicit inductive bias for transfer learning with convolutional
  networks.
\newblock {\em arXiv preprint arXiv:1802.01483}, 2018.

\end{thebibliography}

\end{document}